\documentclass{article}
\usepackage{spconf,amsmath,graphicx}

\title{Unsupervised motion representation enhanced network for action recognition}
\name{Xiaohang Yang, Lingtong Kong, Jie Yang}
\address{Institute of Image Processing and Pattern Recognition\\Shanghai Jiao Tong University, China}
\begin{document}
%
\maketitle
\begin{abstract}
Learning reliable motion representation between consecutive frames, such as optical flow, has proven to have great promotion to video understanding. However, the TV-L1 method, an effective optical flow solver, is time-consuming and expensive in storage for caching the extracted optical flow. To fill the gap, we propose UF-TSN, a novel end-to-end action recognition approach enhanced with an embedded lightweight unsupervised optical flow estimator. UF-TSN estimates motion cues from adjacent frames in a coarse-to-fine manner and focuses on small displacement for each level by extracting pyramid of feature and warping one to the other according to the estimated flow of the last level. Due to the lack of labeled motion for action datasets, we constrain the flow prediction with multi-scale photometric consistency and edge-aware smoothness. Compared with state-of-the-art unsupervised motion representation learning methods, our model achieves better accuracy while maintaining efficiency, which is competitive with some supervised or more complicated approaches.
\end{abstract}
\section{Introduction}
\label{sec:intro}

Convolution Neural Networks (CNNs) have made great progress in multiple image-based tasks, especially in fields like image classification, object detection and segmentation. With the help of massive detailed annotated data, networks can model complex spatial feature and make prediction in an end-to-end manner. However, due to the difficulty of labeling video data and learning spatio-temporal information, human action recognition, although has attracted much attention from researchers over the last few years, still remains a challenging task. We believe that temporal information is the main obstacle of video analysis, therefore, learning effective motion representation is our focus.


\textbf{Problem.} Motion between frames has been proven to play a key role in action recognition task \cite{twostream, twostreamconvfusion}. While some research has been conducted to investigate how to exploit motion cues directly from video sequences \cite{C3D, I3D, P3D, S3D, r2plus1d}, hand-craft optical flow is still effective and widely used for motion description in video understanding. The two-stream model and its variants apply optical flow as one input modality and achieve remarkable performance \cite{twostream, twostreamconvfusion, I3D, hiddentwostream}.

Optical flow could be calculated by optimization methods \cite{tvl1} or pre-trained networks \cite{flownet, flownet2, pwcnet}. Most existing methods take the pre-generated and cached TV-L1 optical flow as the input of motion stream. Compared with other approaches on action datasets, TV-L1 optical flow contains more accurate edges and less noise in non-boundary regions, which makes TV-L1 optical flow the prominent choice of explicit motion representation. However, the pre-processing of optical flow is time-consuming and expensive in storage, which destroys the efficiency in real-world application. Mainstream approaches calculate optical flow and cache them for every two consecutive frames in advance. Although there are CNN-based approaches meeting the requirements of efficiency and accuracy on optical flow benchmarks \cite{flownet, flownet2, pwcnet}, applying these methods directly with their pre-trained models is not as good as TV-L1 methods \cite{hiddentwostream, pclnet}, and it is hard to train dense optical flow on action recognition datesets like HMDB51 \cite{hmdb} and UCF101 \cite{ucf}, due to the difficulty of obtaining reliable ground truth label in real-world situations.

\textbf{Approach.} Motivated by recent advances in unsupervised optical flow estimation \cite{unflow, ddflow}, we address the problem by proposing UF-TSN, a novel lightweight and effective temporal action recognition framework with unsupervised optical flow estimation embedded. More specifically, we replace the temporal stream with an end-to-end two-stage network. The new temporal stream decouples the motion-based action classification into frame-to-motion stage and motion-to-class stage by first exploiting the motion dynamics explicitly without supervision, and then processing the estimated optical flow with a classifier. The brightness consistency based unsupervised loss and edge-aware smoothness term would guide the dense optical flow learning from adjacent image pairs in the first stage, and thus the second stage could combine with different classifiers flexibly.

\begin{figure*}
	\begin{minipage}[b]{.74\linewidth}
		\centering
		\centerline{\includegraphics[width=13.2cm]{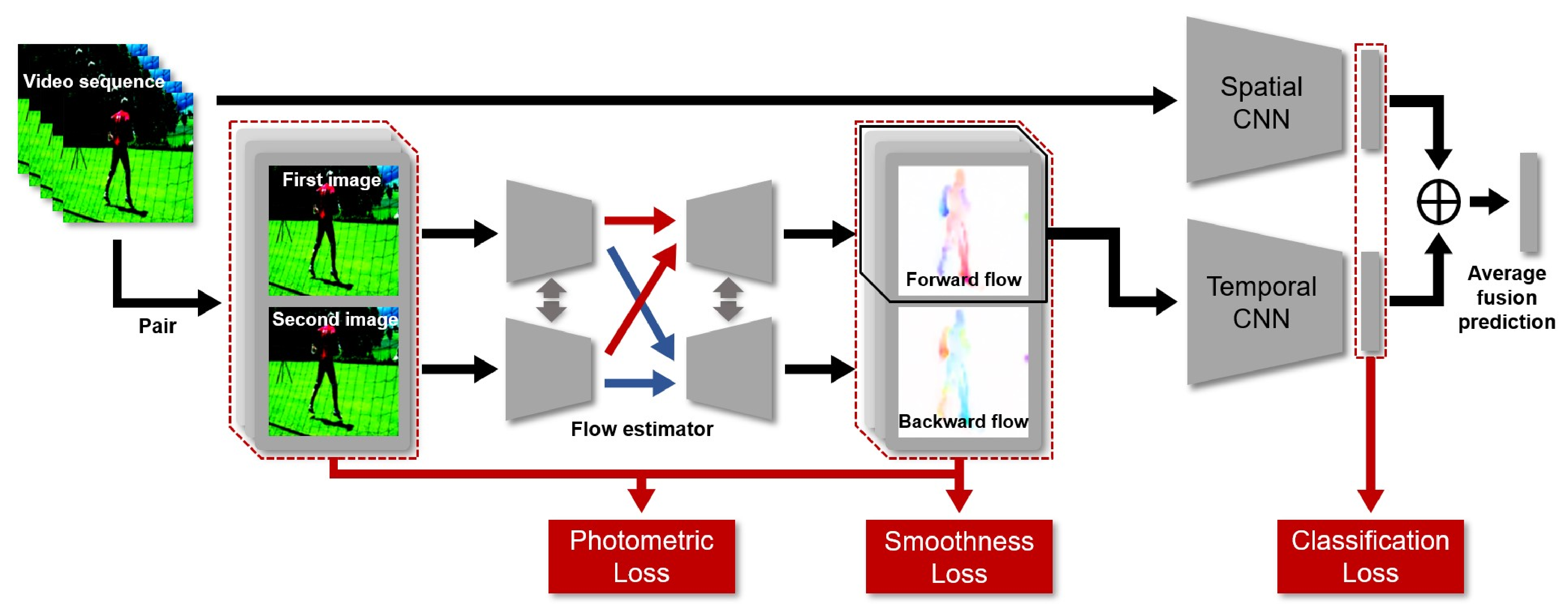}}
		\vspace{-0.1cm}
		\centerline{(a)}\medskip
	\end{minipage}
	\hfill
	\begin{minipage}[b]{0.26\linewidth}
		\centering
		\centerline{\includegraphics[width=3.6cm]{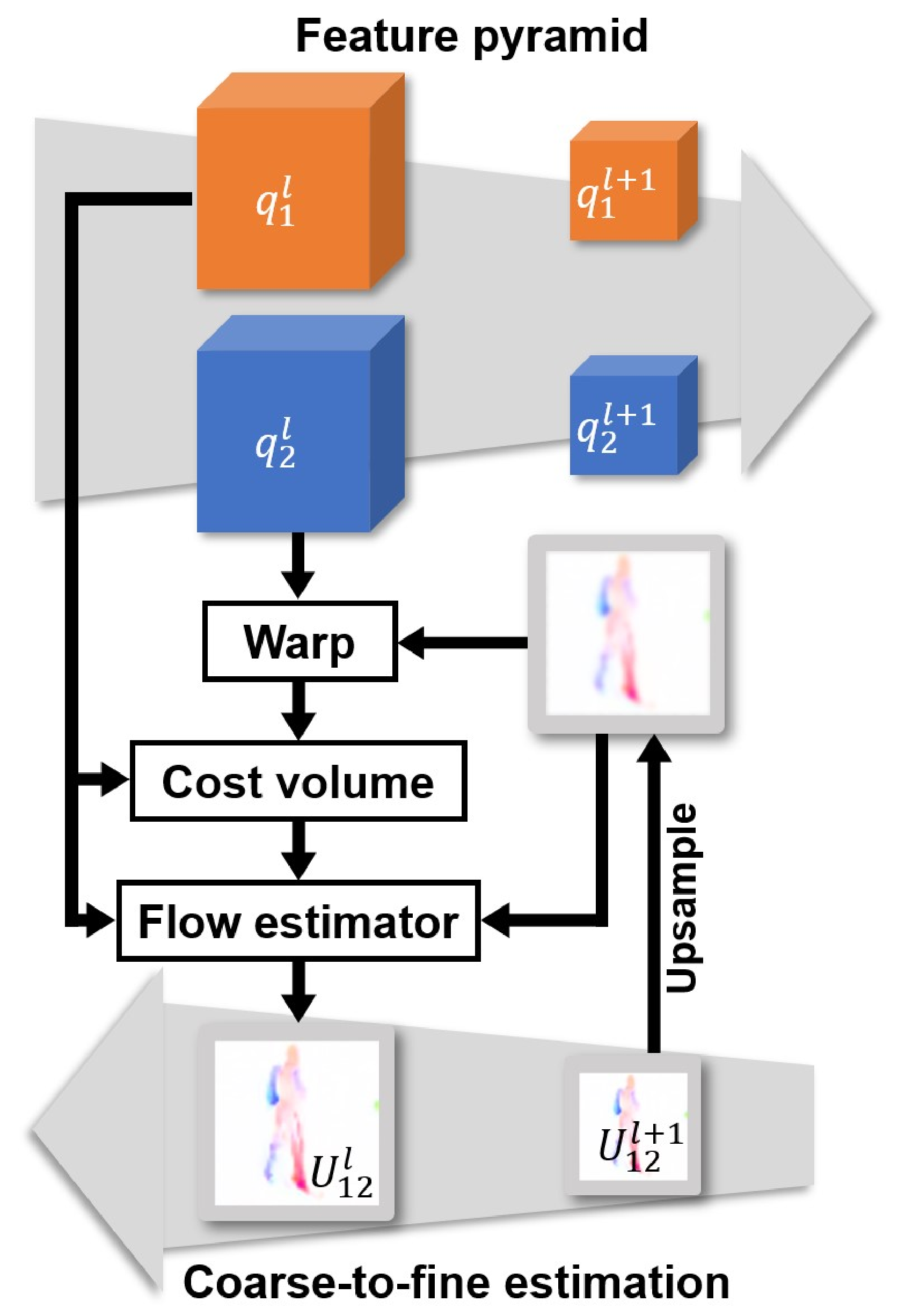}}
		\vspace{-0.1cm}
		\centerline{(b)}\medskip
	\end{minipage}
	\vspace{-1.1cm}
	\caption{(a) Our UF-TSN framework for action recognition. (b) Illustration of the process for estimating flow at one level.}
	\label{image1}
	\vspace{-0.5cm}
\end{figure*}

Furthermore, predicting motion directly from consecutive frames is difficult, thus our UF-TSN estimates flow in a coarse-to-fine manner. We extract feature pyramids to reduce the motion scale and start the exploitation from the top level. Then at each level we warp one frame towards the other according to upsampled optical flow from previous level, UF-TSN therefore only needs to estimate a residual flow.

Finally, we combine the flow estimation and classification together and achieve end-to-end training, we design photometric loss and edge-aware smoothness loss for each level and Cross Entropy Loss to constrain the learning process.

Our method, in summary, is designed to directly work on RGB frames and learns spatial and temporal feature separately before fusion, the contributions are mainly in three aspects: (1) we propose a novel framework based on feature pyramid, warping operation and cost volume to extract motion cues for action classification; (2) we adopt multi-scale unsupervised losses to constrain the flow estimation; (3) UF-TSN achieves better accuracy than similar methods while maintaining great efficiency of over 130 fps on a single GPU.

\textbf{Related Works.} Recently, a lot of methods have been proposed to exploit motion cues from the temporal dimension, which can be divided into three categories. First, 3D convolution networks \cite{C3D, I3D, P3D, r2plus1d} pay equal attention to the extra time dimension and apply 3D convolution to extract spatio-temporal feature implicitly. These methods achieve high accuracy, however, the computation complexity is much higher than 2D convolution. The second category processes spatial and temporal information separately and takes both RGB frames and optical flow as input \cite{twostream, twostreamconvfusion, tsn}. The third category explores the motion representation of compressed videos, which takes motion vector and residual error in compression techniques as new modalities of additional stream \cite{emvnet, coviar, dmcnet}. As an important modality in the second category introduced above, optical flow has attracted a lot of attention from researchers, including traditional optimization methods \cite{tvl1}, supervised methods trained with pixel-wise labels \cite{tvl1, flownet, flownet2, pwcnet, fdflownet, tvnet} and unsupervised approaches trained by brightness consistency\cite{ddflow, pclnet}. 








\section{Method}
\label{sec:pagestyle}

Our UF-TSN, as illustrated in Fig.\ref{image1}(a), consists of two forward stream, the spatial one encodes the semantic information including human poses, objects and backgrounds, while the temporal stream, extracts explicit motion characteristics, which are processed by classifier in the second stage. The final prediction is the average fusion of the scores from these two streams.

\subsection{Unsupervised Motion Estimator}

Reliable motion is one of the requirements for accurate action classification. Recent researches on optical flow learning suggest that, (1) feeding correlation maps rather than RGB image pairs into the network is more helpful for motion cues extraction \cite{flownet, flownet2}, (2) small motion is more easier for correspondence matching, and large displacement could be narrowed with feature pyramids \cite{pwcnet} and warping by coarse optical flow \cite{stn}. We follow these perception to build a more practical architecture for action recognition task, thus we design a lightweight correlation network based on feature pyramids.

Specifically, given two consecutive frames $\{I_1, I_2\}$ from a video sequences $\mathcal{I}$, we aim to train a network with learnable parameters $\Theta$ to predict the forward optical flow $U_{12}$,
\begin{equation}
	U_{12} = f(I_1, I_2, \Theta),
\end{equation}
for later classification, while the backward flow $U_{21}$ is also estimated as data augmentation during training.

\textbf{Coarse-To-Fine Architecture.} We first apply a shallow feature extractor to generate two feature pyramids $q_i^l, i \!\in\! \{1,2\}, l \!\in\! \{0, 1, 2, 3, 4 \}$, where $q_i^0 \!=\! I_i$. Large displacement at $(l\!-\!\!1)$-th level can be reduced by the factor of $2$ at $l$-th level, therefore, we can estimate flow from top level with motion within a few pixels. With optical flow interpolated from previous level, each layer could warp the target feature map towards the source one, and repeat the flow estimation process. The result flow of $l$-level is the sum of the upsampled flow from $(l\!+\!\!1)$-level and predicted residual flow of current level. Specially, the top level initialize the flow with zero.

As shown in Fig.\ref{image1}(b), at $l$-level, forward flow $U_{12}^l$ is constructed based on the feature map of $I_1$ and the warped feature map of $I_2$ according to the upsampled optical flow of last level $\textup{up}(U_{12}^{l+1})$, where the warping operation which is implemented with bilinear interpolation and formulated as:
\begin{equation}
	\hat{q}_2^l(\mathbf{x}) = q_2^l(\mathbf{x} + \textup{up}(U_{12}^{l+1})(\mathbf{x})),
\end{equation}
where $\mathbf{x}$ denotes coordinates of pixels and $\textup{up}(\cdot)$ represents the bilinear interpolation based upsample function.

\textbf{Cost Volume Based Flow Estimation.} Then, we calculate the cost volume between two pixels $\mathbf{x}_1$, $\mathbf{x}_2$ from two feature maps $q_1^l$, $\hat{q}_2^l$ within a local search square :
\begin{equation}
	\textup{CostVolume}^l(\mathbf{x}_1, \mathbf{x}_2) = \dfrac{1}{N} {q_1^l(\mathbf{x}_1)}^\mathsf{T} \hat{q}_2^l(\mathbf{x}_2),
\end{equation}
where $N$ is the length of $q_1^l(\mathbf{x}_1)$.

Next, as flow estimation of the higher level can serve as a coarse estimate for subsequent level, and the next level could predict residual flow to update a more sophisticated one, we construct a network $G^l(\cdot)$ to predict the residual flow $u_{12}^l$ and the output flow $U_{12}^l$ of $l$-level:
\begin{align}
	u_{12}^l &= G^l\left(\textup{cat}(q_1^l, \textup{up}(U_{12}^{l+1}), \textup{CostVolume}^l)\right), \\
	U_{12}^l &= \textup{up}(U_{12}^{l-1}) + u_{12}^l,
\end{align}
the $\textup{cat}$ represents the concatenation operation.

We take the $U_{12}^0$ as the input of the recognition network, in our motion stream we take ResNet-18 as our backbone.

\subsection{Unsupervised Loss}

Due to lack of supervision for action recognition datasets, we adopt photometric loss to constrain the training. Given two consecutive frames $I_1, I_2$, we evaluate the loss of the reconstructed optical flow $U_{12}$ by measuring the pixel-level similarity of $I_1$ and warped $\hat{I}_2$ in a multi-scale manner. We take the forward consistency loss of $l$-level as example.

\textbf{Photometric Loss.} We use $\mathcal{L}_1$-norm and census loss \cite{unflow} as our photometric loss to guide the training at pixel level and patch level respectively.


\begin{equation}
	\vspace{-0.2cm}
	\mathcal{L}_{\textup{diff}} = \sum_{i,j}^M \left\|I_1^l(i, j) - \hat{I}_2^l(i, j)\right\|_1,
\end{equation}


\begin{equation}
	\mathcal{L}_{\textup{census}} = \sum_{i, j}^M d(c_1^l(i, j), c_2^l(i, j)),
\end{equation}
where $M$ is the pixel number of the image, $p_1^l (i,j)$ is the small patch of $3 \times 3$ from $I_1^l$ at $(i,j)$, $d(\cdot)$ is the Hamming distance, and $c_1^l(i,j)$ is the census code at $(i,j)$.

\textbf{Smoothness Loss.} Clear boundary and smooth motion field inside the moving parts is crucial for extraction of high-level movement information. We use image gradient as the weight of the smoothness loss to constrain the flow changing in non-boundary area:
\begin{equation}
	\mathcal{L}_{\textup{smooth}} = \sum_{d=x, y} \sum_{i, j}^M \left\|\nabla_d U_{12}^l\right\|_1 e^{-\left(\left\|\nabla_d I_1^l\right\|_1 \right)}.
\end{equation}

The overall loss function is formulated as:
\begin{equation}
	\mathcal{L} = \frac{1}{M} \left( \lambda_1 \mathcal{L}_{\textup{diff}} + \lambda_2 \mathcal{L}_{\textup{census}} + \lambda_3 \mathcal{L}_{\textup{smooth}} \right).
\end{equation}



\section{Experiment}
\label{sec:typestyle}

\subsection{Datasets}

We investigate the performance of our model on two popular datasets of realistic action videos, UCF101 \cite{ucf} and HMDB51 \cite{hmdb}. UCF101 collects 13320 videos from 101 action categories and HMDB51 contains 6766 videos from 51 classes. We follow the 3 public train/test splits and evaluate top-1 accuracy on video level \cite{ucf, hmdb}.

\subsection{Implementation Details}
\label{implementation}

\textbf{Training.} We use ResNet-152 for the spatial stream and use ResNet-18 following the flow estimator to predict categories in consideration of the trade-off between efficiency and accuracy. Both streams take RGB frames as input, while the motion network stacks a video sequences of 10 frames. For spatial stream, we initialize the flow estimator randomly and the classifier with the parameter pre-trained on ImageNet. Due to the difficulty of training a two-stage model, we split the training process into three stage. We first train the flow estimator to obtain a stable and reliable estimation with $\lambda_1=0.2, \lambda_2=1.0, \lambda_3=50.0$. Then, we detach the flow to warm up the classifier with Cross Entropy Loss. Finally, we fine-tune the two-stage network end-to-end.

\textbf{Testing.} For each video, 25 input is sampled uniformly for both stream with no augmentation. We follow CoViAR \cite{coviar} and DMC-Net \cite{dmcnet} to make an average fusion of the scores from two modalities, which collects 25 scores for each video and modality, then calculate an average score to predict video-level classification.

\begin{table}[ht]
	\vspace{-0.1cm}
	\caption{Comparison with state-of-the-arts on UCF101 and HMDB51. All methods are divided into three categories according to the difference of input modalities. The bottom section is combination of two-stream and preprocess optical flow methods, whose accuracy rates are from our implementation, with the same classifier, training strategy. The underlined methods extract optical flow without supervision.}
	\centering
	\begin{tabular}{p{4.61cm}p{1.28cm}p{1.38cm}}
		\hline
		Methods                   & UCF101  & HMDB51 \\
		\hline
		\textbf{Compressed Video} & & \\
		CoViAR \cite{coviar}      & 90.4   & 59.1      \\
		DMC-Net \cite{dmcnet}     & 90.9   & 62.8    \\
		\hline
		\textbf{RGB only} & & \\
		\underline{PCLNet} \cite{pclnet} & 82.8  & 53.5 \\
		\underline{ActionFlowNet} \cite{actionflownet}  & 83.9   & 56.4 \\
		\underline{Hidden two-stream (VGG16)} \cite{hiddentwostream}  & 90.3  & 60.5 \\
		TV-Net \cite{tvnet}  & 94.5   & 71.0 \\
		\underline{Our UF-TSN with ResNet-18}  & 92.2   & 64.4 \\
		\hline
		\textbf{RGB + Optical Flow}  &  &  \\
		Two-stream \cite{twostream}  & 88.0  & 59.4 \\
		Two-stream Fusion \cite{twostreamconvfusion}  & 92.5   & 65.4 \\
		\hline
	\end{tabular}
	\label{table1}
	\vspace{-0.6cm}
\end{table}

\begin{table}[ht]
	\caption{Performances of other optical flow approaches and ablation study.}
	\centering
	\begin{tabular}{p{4.61cm}p{1.28cm}p{1.38cm}}
		\hline
		Methods                   & UCF101  & HMDB51 \\
		\hline
		TV-L1 \cite{tvl1}  & 92.2  & 64.5 \\
		PWC-Net \cite{pwcnet}  & 91.3  & 62.8 \\
		\hline
		w/o cost-volume \& multi-scale  & 90.4  & 60.9 \\
		w/o multi-scale  & 91.2  & 62.5 \\
		\hline
	\end{tabular}
	\label{table3}
	\vspace{-0.6cm}
\end{table}

\subsection{Comparison With State-Of-The-Art Methods}

\begin{figure}[b]
	\vspace{-0.6cm}
	\centering
	\begin{minipage}[b]{1.0\linewidth}
		\centering
		\centerline{\includegraphics[width=8.6cm]{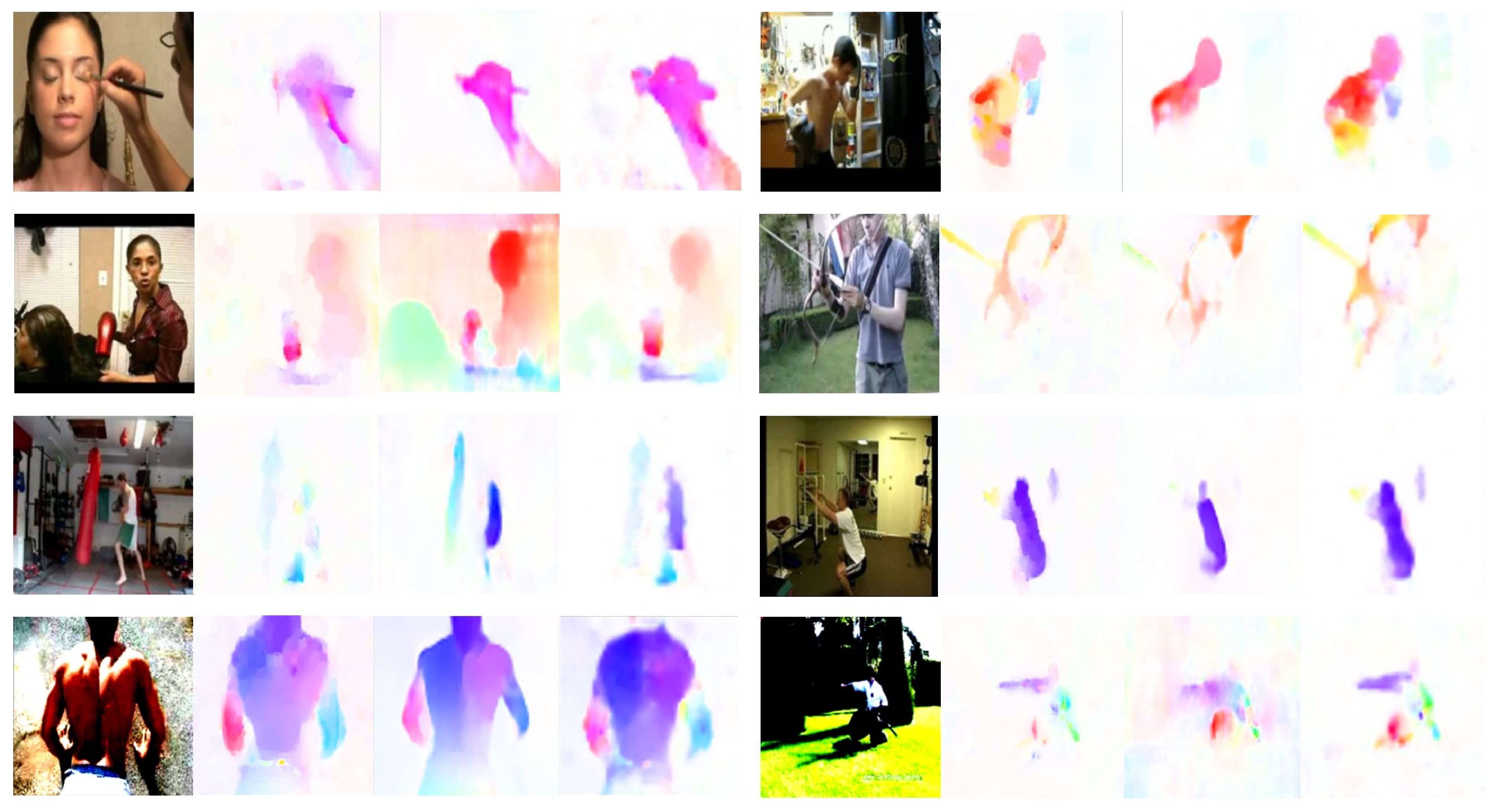}}
	\end{minipage}
	\vspace{-0.8cm}
	\caption{Visualization of optical flow. From left to right: original RGB image, TV-L1, PWC-Net, UF-TSN.}
	\label{image2}
	\vspace{-0.1cm}
\end{figure}

As our UF-TSN takes only RGB data as input and the basic framework is based on two-stream network with average fusion, we compare the accuracy mainly with methods which take only RGB frames or have similar frameworks of two-stream or multi-stream. Table \ref{table1} shows the accuracy on two datasets, compared our model with various methods.

Our UF-TSN outperforms three unsupervised motion representation learning methods, PCL-Net \cite{pclnet}, ActionFlowNet \cite{actionflownet} and Hidden two-stream \cite{hiddentwostream}, under the same or similar performances of the backbone classifiers, suggesting that our motion representation is more appropriate for action recognition tasks. The backbones of ActionFlowNet \cite{actionflownet} and Hidden two-stream \cite{hiddentwostream} are like Encoder-Decoder, which are weaker on distinguishing boundary than our model and result in blurred motion. Our UF-TSN is not as good as TV-Net \cite{tvnet}, which is first trained on optical flow datasets and applies a stronger classifier. Compared with CoViAR \cite{coviar} and DMC-Net \cite{dmcnet}, whose results fuse three and four modalities respectively, our model outperforms these two methods when using ResNet only. We argue that MV and Residual are two weak modalities as they decouple the complete motion into two splits. The accuracy of two-stream \cite{twostream} and two-stream fusion \cite{twostreamconvfusion} is displayed in the third block of Table \ref{table1}, and our results is close to the two-stream fusion methods with only average fusion.



In Table \ref{table3}, we replace our optical flow estimation with two pre-generated methods, TV-L1 \cite{tvl1} and PWC-Net \cite{pwcnet}, and give two extra realizations for ablation study. Our methods achieve a closer performance to the TV-L1 optical flow than other unsupervised motion representation learning methods like ActionFlowNet \cite{actionflownet} and Hidden two-stream \cite{hiddentwostream}. Eliminating feature pyramid and cost volume for ablation study, our UF-TSN results in a similar structure with Hidden two-stream \cite{hiddentwostream}, and the drop in accuracy shows the effectiveness of feature pyramid and cost volume on motion learning.

\begin{figure}
	\centering
	\begin{minipage}[b]{1.0\linewidth}
		\centering
		\centerline{\includegraphics[width=8.6cm]{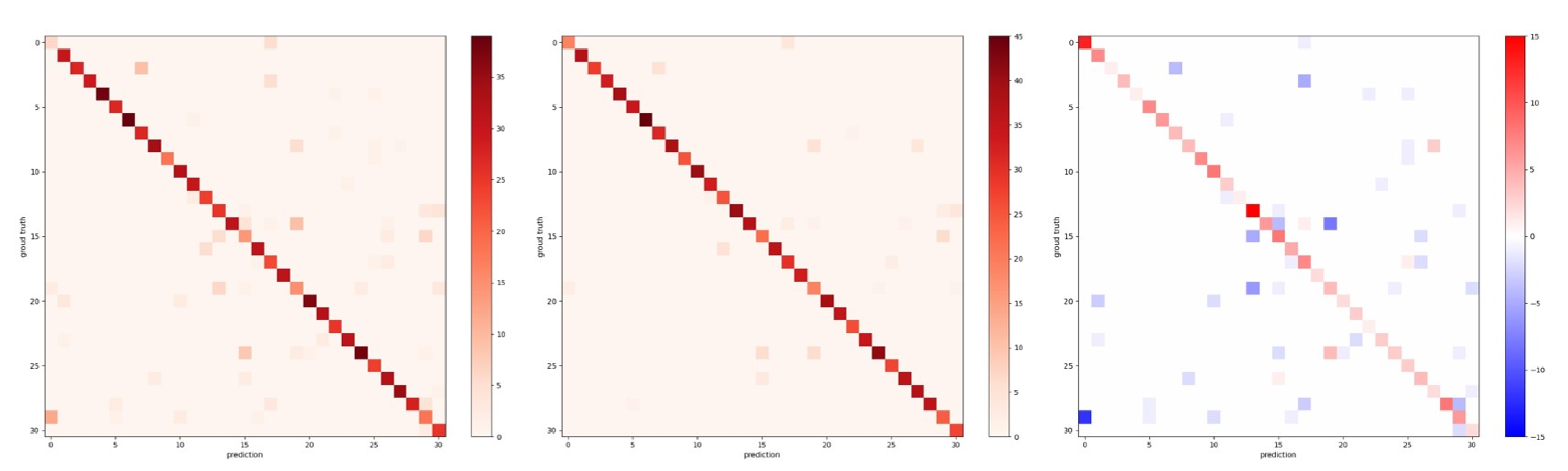}}
	\end{minipage}
	\vspace{-0.8cm}
	\caption{Confusion matrix of part UCF101. From left to right: RGB stream, UF-TSN, promotion with our methods.}
	\label{image3}
\end{figure}

The visualization of estimated flow is shown in Fig.\ref{image2}. We compare UF-TSN with TV-L1 \cite{tvl1} and PWC-Net \cite{pwcnet} on UCF101 and HMDB51. Our method achieves competitive performance with TV-L1 \cite{tvl1}, which is the best result of unsupervised motion learning methods. Displayed in Fig.\ref{image3} is part of the confusion matrix on UCF101, the red color in the right graph means the improvement our UF-TSN makes over some confusing categories for spatial stream, especially the action that is not restricted by the scene.

\textbf{Computational Efficiency Comparison.} We compare the inference speed with other motion representation learning methods. The measured time consists of pre-processing and forward propagation. Inference speed and parameters can be found in Table \ref{table2}, UF-TSN is significantly faster than most methods, if increasing the batch size, the Frames-per-Second (FPS) can improve to over 300 on a single RTX2070 GPU.

\begin{table}[]
	\vspace{-0.4cm}
	\caption{Evaluation of efficiency for the motion stream. Parameter take only motion estimation into consideration. Time is calculated from RGB images to prediction.}
	\centering
	\begin{tabular}{p{4.50cm}p{1.68cm}p{1.0cm}}
		\hline
		& Params (M)  & FPS \\
		\hline
		PCLNet \cite{pclnet} & 11.9  & 19.2 \\
		ActionFlowNet \cite{actionflownet}  & -  & 200 \\
		Hidden Two-stream \cite{hiddentwostream}  & -  & 48.5 \\
		TV-Net \cite{tvnet}  & 0.1  & 12.0 \\
		UF-TSN motion stream  & 5.7  & 131.6 \\
		\hline
	\end{tabular}
	\label{table2}
	\vspace{-0.6cm}
\end{table}





\section{Conclusion}

In this paper, we propose UF-TSN, a novel unsupervised motion representation learning framework for action recognition. UF-TSN applies feature pyramid and warping operation to reduce large displacement and estimates flow based on cost volume from coarse to fine, and then we constrain the prediction with image reconstruction and edge-aware smoothness losses in a multi-scale manner. Classification accuracy and visual instances of UF-TSN on two benchmark datasets have quantitatively and qualitatively demonstrated the competitive performance with TV-L1, which maintains the efficiency at the same time.

\vfill\pagebreak


\bibliographystyle{IEEEbib}
\bibliography{strings,refs}

\end{document}